\documentclass[letterpaper, 10 pt, conference]{ieeeconf}  %

\IEEEoverridecommandlockouts                           

\usepackage{times}
\usepackage{microtype}
\usepackage{epsfig}
\usepackage[table,xcdraw]{xcolor}
\usepackage{caption}
\usepackage{float}
\usepackage{placeins}
\usepackage{color, colortbl}
\usepackage{stfloats}
\usepackage{tabularx}
\usepackage{xstring}
\usepackage{multirow}
\usepackage{booktabs}
\usepackage{xspace}
\usepackage{url}
\usepackage{subcaption}
\usepackage{xcolor}
\usepackage[hang,flushmargin]{footmisc}
\usepackage{algorithm}
\usepackage[noend]{algpseudocode}
\usepackage{amsfonts}
\usepackage{amsmath}
\usepackage{bbm}
\usepackage{hyperref}
\usepackage{amssymb}
\usepackage{dsfont}
\usepackage{bm}
\usepackage{soul}
\usepackage{lipsum}

\makeatletter
\DeclareRobustCommand\onedot{\futurelet\@let@token\@onedot}
\def\@onedot{\ifx\@let@token.\else.\null\fi\xspace}

\def\etal{\emph{et al}\onedot}
\makeatother

\newcommand{\R}[1]{{%
    \textbf{%
        \ifstrequal{#1}{1}{\textcolor{red}{R#1}}{%
        \ifstrequal{#1}{2}{\textcolor{blue}{R#1}}{%
        \ifstrequal{#1}{3}{\textcolor{magenta}{R#1}}{%
        \ifstrequal{#1}{4}{\textcolor{teal}{R#1}}{%
                           \textcolor{cyan}{R#1}%
        }}}}%
    }%
}}

\newcounter{taborder}

\title{\LARGE \bf
ViViDex: Learning Vision-based Dexterous Manipulation\\ from Human Videos
}

\author{Zerui Chen$^1$, Shizhe Chen$^1$, Etienne Arlaud$^1$, Ivan Laptev$^{2}$ and Cordelia Schmid$^1$%
\thanks{$^1$Inria, \'Ecole normale sup\'erieure, CNRS, PSL Research University, 75005, Paris, France. {\tt\small \{firstname.lastname\}@inria.fr}}\\
\thanks{$^2$Mohamed bin Zayed University of Artificial Intelligence, Abu Dhabi, United Arab Emirates. {\tt\small \{firstname.lastname\}@mbzuai.ac.ae}}\\
}

\begin{document}

\bstctlcite{IEEEexample:BSTcontrol}

\maketitle
\thispagestyle{empty}
\pagestyle{empty}

\begin{abstract}
In this work, we aim to learn a unified vision-based policy for multi-fingered robot hands to manipulate a variety of objects in diverse poses. Though prior work has shown benefits of using human videos for policy learning, performance gains have been limited by the noise in estimated trajectories. Moreover, reliance on privileged object information such as ground-truth object states further limits the applicability in realistic scenarios. To address these limitations, we propose a new framework ViViDex to improve vision-based policy learning from human videos. It first uses reinforcement learning with trajectory guided rewards to train state-based policies for each video, obtaining both visually natural and physically plausible trajectories from the video. We then rollout successful episodes from state-based policies and train a unified visual policy without using any privileged information. We propose coordinate transformation to further enhance the visual point cloud representation, and compare behavior cloning and diffusion policy for the visual policy training. Experiments both in simulation and on the real robot demonstrate that ViViDex outperforms state-of-the-art approaches on three dexterous manipulation tasks. Project website: \url{zerchen.github.io/projects/vividex.html}
\end{abstract}
\begin{figure*}[ht]
  \centering
\includegraphics[width=0.95\linewidth]
{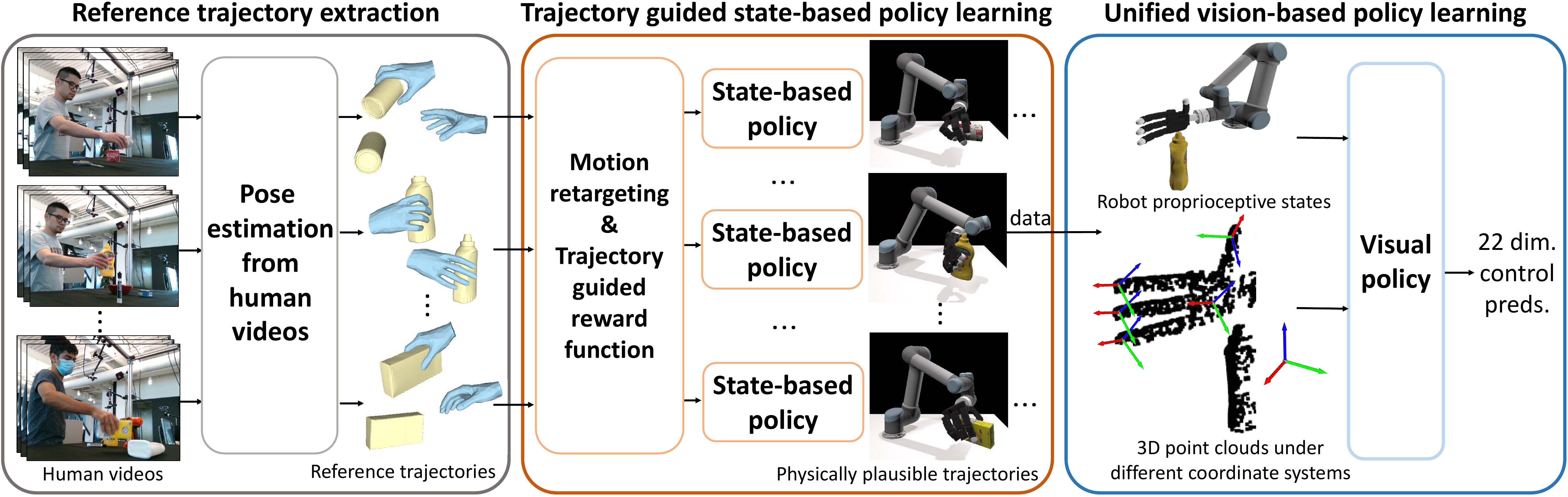}
  \vspace{-0.1cm}
  \caption{The overall framework of our method for learning dexterous manipulation skills from human videos. It consists of three steps: extraction of reference trajectories from human videos, trajectory-guided state-based policy learning using RL, and vision-based policy learning using either the behavior cloning or the 3D diffusion policy. 
}
  \label{fig:overview}
  \vspace{-0.5cm}
\end{figure*}

\section{Introduction}
\label{sec:intro}
People possess a remarkable ability to manipulate objects effortlessly with their hands, guided by visual perception.
Despite significant progress, replicating the dexterity of human hands with multi-fingered robot hands remains challenging~\cite{james2020rlbench,guhur2023instruction,driess2023palm,chen23polarnet,mason2018toward}.
This is largely due to the complexity of translating visual signals into the high-dimensional control commands required for dexterous manipulation.

Recent developments in deep learning (DL)~\cite{lecun2015deep}
and reinforcement learning (RL)~\cite{kaelbling1996reinforcement,sutton2018reinforcement} have enabled significant progress in learning-based algorithms for dexterous manipulation such as in-hand rotation~\cite{qi2022hand,yin2023rotating,qi2023general}, solving Rubik's cube~\cite{akkaya2019solving} and playing the piano~\cite{xu2022towards,zakka2023robopianist}. However, it proves challenging to train RL policies as RL training heavily relies on intricate reward engineering~\cite{sutton2018reinforcement} and extensive computational resources~\cite{akkaya2019solving}.
Furthermore, unnatural behaviors with high rewards can emerge from RL training~\cite{merel2017learning}.
To address the limitations of RL, some previous works~\cite{rajeswaran2017learning,qin2023anyteleop,arunachalam2023dexterous,arunachalam2023holo,wang2024dexcap,yang2024ace} have turned to imitation learning using robot demonstrations collected by teleoperation. While these approaches enhance training efficiency, they demand substantial human efforts to collect diverse robot demonstrations and are, hence, difficult to scale.  

As human videos capturing hands manipulating objects~\cite{YangCVPR2022OakInk,liu2022hoi4d} are abundant and easy to acquire, 
there is a growing trend in robotics research to utilize human video demonstrations for learning dexterous manipulation skills~\cite{qin2022dexmv,mandikal2022dexvip,liu2023dexrepnet,shaw2022videodex,dasari2023pgdm}. 
For example, DexMV~\cite{qin2022dexmv} proposes to extract robot and object poses from human videos, and then employs the data to accelerate RL training via demonstration augmented policy gradient~\cite{rajeswaran2017learning}. 
However, since the extracted robot and object poses are noisy, the method needs hundreds of human videos to learn the manipulation of a single object and requires reward engineering for different tasks. Moreover, existing methods~\cite{qin2022dexmv,liu2023dexrepnet,wu2023learning} often leverage privileged information of objects in policy learning such as ground-truth object CAD models and object poses, which are non-trivial to obtain from visual sensor data in real-world scenarios. 

In this work, we propose ViViDex, a new framework making use of human \textbf{Vi}deos for \textbf{Vi}sion-based \textbf{Dex}terous manipulation.
ViViDex consists of three modules as illustrated in Figure~\ref{fig:overview}.
We first obtain human hand and object trajectories from video demonstrations~\cite{qin2022dexmv}.
Though such trajectories are noisy and not directly usable for robot control, they provide examples of natural hand-object interactions.
Our second module then refines the reference trajectories to be physically plausible. We train a state-based policy with RL for each reference trajectory. A novel trajectory-guided reward is proposed to keep pose similarity in the reference trajectory while solving the task. We also augment trajectories during training to generalize to a broader range of object poses beyond the pose in the reference trajectory.
Finally, we rollout episodes from optimized state-based policies and distill successful episodes to a unified vision-based policy using no privileged object information.
To this end, we propose to improve visual representation by transforming input 3D point clouds into hand-centered coordinate systems.
We evaluate ViViDex on three dexterous manipulation tasks \emph{relocation}, \emph{pour} and \emph{placing inside} in simulation and demonstrate significant improvements over the state-of-the-art DexMV~\cite{qin2022dexmv} while using significantly less human demonstrations. Our visual policy also achieves good performance both in simulation and on a real robot for both seen and unseen objects.

To summarize, our contributions are three-fold:

\noindent
\textbf{$\bullet$} We introduce a new framework ViViDex to learn vision-based dexterous manipulation policy from human videos.

\noindent
\textbf{$\bullet$} We improve the RL rewards for the state-based policy by imitating trajectories extracted from human videos and propose a novel model architecture for the vision-based policy.

\noindent
\textbf{$\bullet$} Extensive simulation and real robot experiments demonstrate the effectiveness of our proposed ViViDex approach.
\section{Related Work}
\label{sec:related}

\noindent \textbf{Dexterous manipulation}. 
Multi-fingered robotic hands enable robots to perform delicate manipulation operations on objects.
Previous work addressed dexterous manipulation by trajectory optimization~\cite{bicchi2000robotic,han1998dextrous,rus1999hand,mordatch2012contact,wu2022learning,yang2024contactsdf,zhu2023diff} and data-driven learning methods~\cite{rajeswaran2017learning,qin2022dexmv,mandikal2022dexvip,wu2023learning,nairr3m,xu2023unidexgrasp,chen2022system,bao2023dexart,lin2024learning,han2023utility,wang2024cyberdemo}. Trajectory optimization methods usually require well-defined dynamic models for the robot and the manipulated object, which are sometimes hard to obtain in practice. Data-driven methods instead directly train neural policies given robot data.
Most of the works~\cite{yuan2024robot,chen2023visual,rajeswaran2017learning,wu2023learning,xu2023unidexgrasp,wan2023unidexgrasp++,nagabandi2020deep,hansen2022modem} use RL algorithms to train policies. Nagabandi~\etal~\cite{nagabandi2020deep} and Hansen~\etal~\cite{hansen2022modem} propose to improve the sample efficiency in RL by jointly learning the dynamic model and the control policy. To accelerate the convergence of RL, Rajeswaran \etal~\cite{rajeswaran2017learning} introduce demonstration augmented policy gradients (DAPG) and train RL with expert demonstrations. Instead of manually collecting expert demonstrations, ILAD~\cite{wu2023learning} derives demonstration data via an affordance model~\cite{jiang2021hand} and trajectory optimization~\cite{rubinstein1999cross}. Xu \etal~\cite{xu2023unidexgrasp} and Wan \etal~\cite{wan2023unidexgrasp++} decompose the dexterous grasping into two sub-tasks: the grasp proposal generation and the goal-conditioned policy training and largely improve the generalization of the learned policy. However, RL is data-inefficient, requires well-designed rewards, and often results in unrealistic object manipulation. 
In this work, we leverage human videos to improve RL training of state-based policies and then train a unified visual policy 
that enables object manipulation given 3D point cloud inputs.

\noindent \textbf{Learning robotic manipulation from human videos}. Human videos naturally provide rich human hands interactions with diverse objects~\cite{chao2021dexycb,hasson2019learning,chen2022alignsdf,chen2023gsdf,ye2022s}. Equipping robots with the capacity to acquire manipulation skills by simply watching human videos has been an attractive direction~\cite{smith2019avid,zakka2021xirl,ebert2021bridge,schmeckpeper2020reinforcement,shao2021concept2robot}. 
One line of works~\cite{nairr3m,xiao2022masked,radosavovic2023real,Ze2023HInDex,chane2023learning} 
aims to learn generic visual representations from large-scale human video data~\cite{grauman2022ego4d} and then makes use of such pre-trained representations to learn control policies.
Another line of works~\cite{chen2021learning,alakuijala2023learning,xiong2021learning} focuses on learning reward functions from human videos. Chen~\etal~\cite{chen2021learning} introduce DVD, a domain-agnostic video discriminator, to learn multi-task reward functions. Alakuijala~\etal~\cite{alakuijala2023learning} propose a task-agnostic reward function by using unlabeled human videos. 
Some other works~\cite{qin2022dexmv,mandikal2022dexvip,liu2023dexrepnet,shaw2022videodex,mandikal2021learning} explore how to explicitly leverage human videos to facilitate learning dexterous manipulation skills. Mandikal \etal \cite{mandikal2022dexvip,mandikal2021learning} propose to utilize object affordances priors and human grasping priors to improve the dexterous grasping policy. DexMV~\cite{qin2022dexmv} and DexRepNet~\cite{liu2023dexrepnet} extract demonstrations data from human videos and use DAPG algorithm~\cite{rajeswaran2017learning} to launch the RL training with demonstrations. PGDM~\cite{dasari2023pgdm} proposes to initialize the robot hand configuration at the given pre-grasp derived from human motions and enable efficient RL training. Compared to prior work~\cite{qin2022dexmv,liu2023dexrepnet}, our method allows to solve more complex manipulation tasks with fewer videos by using a unified trajectory guided reward learned from videos and automatic trajectory augmentation.
\section{Method}
\label{sec:method}

This work aims to train vision-based dexterous manipulation policies. However, jointly learning the visual representation and the control policy is challenging. Therefore, we first train the state-based policy and then distill the learned policy to a vision-based policy. To train the state-based policy, as the pure RL training is ineffective~\cite{qin2022dexmv,liu2023dexrepnet,wu2023learning}, we extract reference trajectories from human videos and propose trajectory-guided RL to improve the performance. In the following, we first present the reference trajectory generation in Section~\ref{met:traj}. Then, we introduce our state-based policy learning algorithm in Section~\ref{met:state}. Finally, we describe the visual policy and its training in Section~\ref{met:visual}.

\subsection{Reference Trajectory Extraction}
\label{met:traj}
\begin{figure}[t]
  \centering
\includegraphics[width=0.90\linewidth]{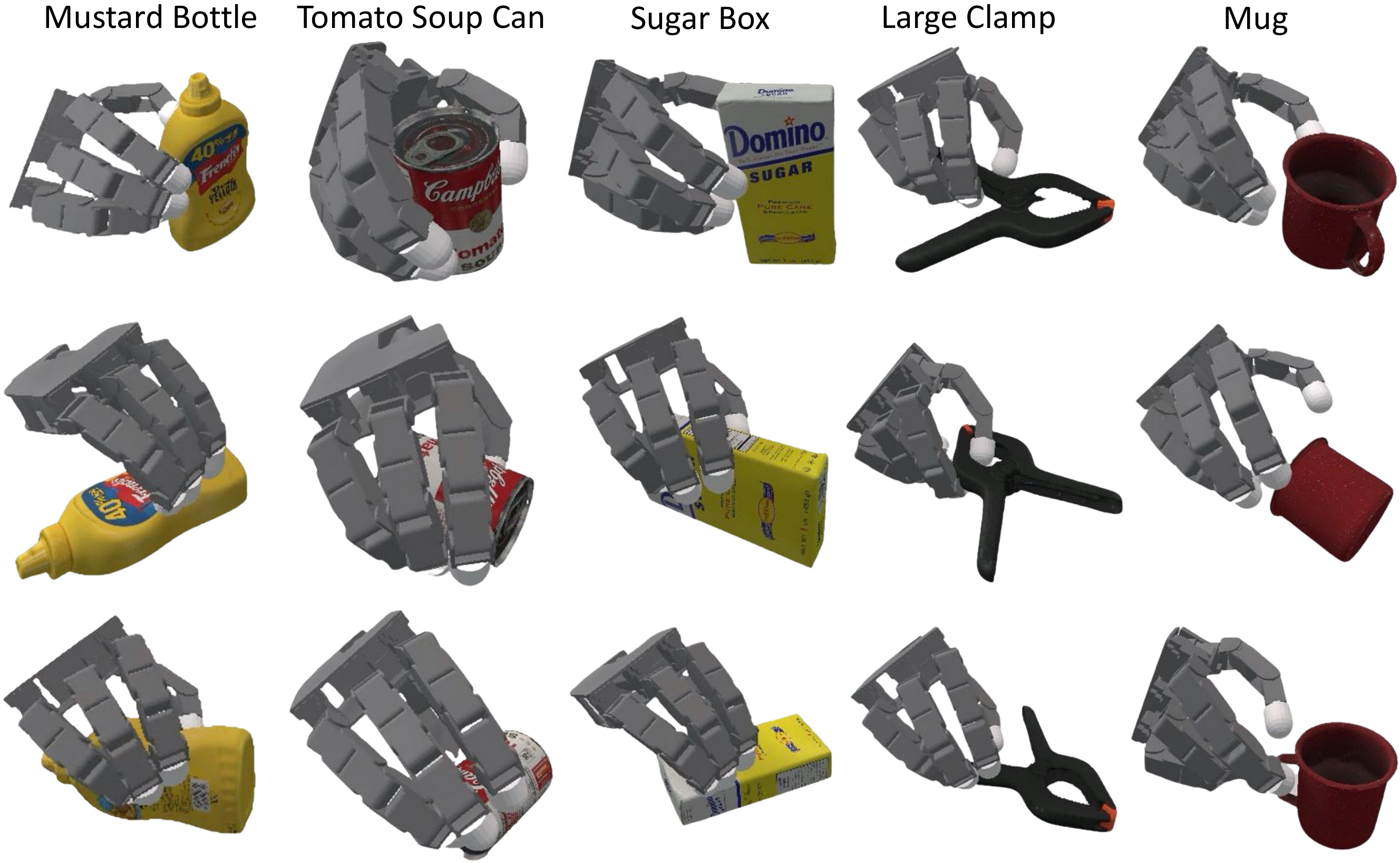}
\vspace{-0.2cm}
  \caption{Motion retargeting results for the Allegro hand and objects under different poses for selected DexYCB videos.}
  \label{fig:object_pose}
  \vspace{-0.6cm}
\end{figure}

To guide our policy training, we extract hand and object poses from video demonstrations.
The poses and shapes of hands are represented by MANO~\cite{MANO:SIGGRAPHASIA:2017} and result in 3D locations of joints $\bm{\psi}_{h} \in \mathbb{R}^{21 \times 3}$. To account for differences in geometry, we retarget the human hand pose to the robot hand pose. Following~\cite{qin2022dexmv,handa2020dexpilot,antotsiou2018task}, we formulate the hand motion retargeting from a video of length $T$ as an optimization problem and define its objective function as:
\vspace{-0.2cm}
\begin{equation}
\begin{split}
\min_{\bm{q}_{r}^{t}} \sum_{t=1}^{T} \Big \|\bm{\hat{x}}_{rj}^{t}(\bm{q}_{r}^{t}) - \bm{{\psi}}_{hj}^{t}\Big \|^{2}_{2} + \alpha \Big \| \bm{q}_{r}^{t} - \bm{q}_{r}^{t-1}\Big \|^{2}_{2},
\end{split}
\label{eq:retarget} 
\end{equation}
where $\bm{q}_{r}^{t}$ represents robot joint rotation angles. $\bm{{\psi}}_{hj}^{t}$ denotes human hand tip and middle phalanx positions.
We solve their robot counterparts $\bm{\hat{x}}_{rj}^{t}$ via forward kinematics according to given ${\bm{q}_{r}^{t}}$. 
The first term aims to minimize the difference between the 3D locations of robot joints to those in the human hand. We also use a regularization term to avoid sudden changes in the robot pose, where we empirically set $\alpha$ to 4e-3 and ${\bm{q}_{r}^{0}}$ to the mean pose of its motion limits. We use the NLopt solver~\cite{stevennlopt} for the optimization and port robot and object motions into the simulator to generate reference trajectories, which are visually plausible but not physically plausible. Figure~\ref{fig:object_pose} illustrates results of the motion retargeting for the Allegro robot hand.

\subsection{State-based Policy}
\label{met:state}

We train a state-based policy with RL to recover physically plausible trajectories, where the above generated reference trajectories are employed in the reward function to guide robot hand and object motions. 
The network architecture for the state-based policy consists of actor and critic MLPs~\cite{schulman2017proximal}. It takes both robot and object states as inputs and predicts the robot control commands. 
In the following, we will describe our reward functions and the trajectory augmentation for RL.

\noindent \textbf{Trajectory-guided reward functions}. During RL training, we propose to divide the reference trajectory into two stages: the pre-grasp stage and the manipulation stage. However, at test time, the policy executes without stage distinctions.

During the pre-grasp stage, the robot hand needs to approach the object without making physical contacts. We require that the robot approaches the object in a similar way as humans and define the following reward function as: 
\vspace{-0.2cm}
\begin{equation}
\begin{split}
R_{p} = \sum_{t=1}^{T_{p}} 10 \cdot {\rm exp}(-10 \cdot \Big \|\bm{{x}}_{rt}^{t}(\bm{q}_{r}^{t}) - \bm{\hat{x}}_{rt}^{t}\Big \|^{2}_{2}), 
\end{split}
\label{eq:pregrasp} 
\end{equation}
where $T_p$ is the length of pre-grasp steps, and $\bm{\hat{x}}_{rt}^{t}$ denotes the robot finger tip positions at the timestep $t$ in the reference trajectory. $\bm{{x}}_{rt}^{t}$ is the current robot finger tip positions.

When the robot successfully reaches its pre-grasp configuration, the episode starts its manipulation stage. In this stage, the robot aims to manipulate the object and bring it to the desired target configuration. Here, we define the reward function to constrain the robot and object motions jointly:
\vspace{-0.2cm}
\begin{equation}
\begin{split}
R_{m}=\sum_{t=T_{p}+1}^{T_{r}} \lambda_{1} R_{m}^{h} + \lambda_{2}R_{m}^{o} + \lambda_{3}\mathds{1}_{\rm cont} + \lambda_{4}\mathds{1}_{\rm lift}, 
\end{split}
\label{eq:mani} 
\end{equation}
where $T_{r}$ is the length of the reference trajectory. The first term $R_{m}^{h}$ constrains the hand motions similar to (\ref{eq:pregrasp}). The second term $R_{m}^{o}={\rm exp}(-\alpha_{1}(\Big \|\bm{x}_{o}^{t} - \bm{\hat{x}}_{o}^{t}\Big \|^{2}_{2} + \alpha_{2}\phi(\bm{\theta}_{o}^{t},~\bm{\hat{\theta}}_{o}^{t})))$ constrains the object motions, where $\bm{x}_{o}^{t}$ and $\bm{\hat{x}}_{o}^{t}$ are the current object position and its reference at timestep $t$. $\phi(\cdot)$ computes the angular distance between the current object orientation with its reference $\bm{\hat{\theta}}_{o}^{t}$. The third term computes the number of finger tips in contact with the object. The forth term assigns bonus points when the object is lifted off the table. We empirically set $\lambda_1=4, \lambda_2=10, \lambda_3=0.5, \alpha_1=50$ and $\alpha_2=0.1$.
Since our reward function is primarily derived from trajectories in videos, it can be applied across various manipulation tasks, reducing the need for task-specific reward engineering as in DexMV~\cite{qin2022dexmv}.

\noindent \textbf{Reference trajectory augmentation}. Though the state-based policy can successfully imitate the hand and object motions from a video, it remains challenging to generalize to different initial object positions, rotations and target positions. 
To make our policy applicable to different initial and target object configurations, we introduce our reference trajectory augmentation strategies during RL training. 
Specifically, we randomly set initial object positions or rotations and then transform the whole reference trajectory accordingly. 
In order to further increase the diversity of target object positions, we augment the original object trajectory through interpolations between the original last object pose and the target pose. We similarly interpolate hand motions and use proposed reward functions to train the state-based policy.

\subsection{The Vision-based Policy}
\label{met:visual}
Though our state-based policy can work well for different manipulation tasks, it requires robot proprioceptive states and object states as inputs. However, reliably estimating object states is often not trivial in practice. To alleviate this issue, we propose a vision-based policy that only takes robot states and 3D scene point clouds as inputs. To generate training data for the visual policy, we rollout the optimized state-based policy and generate diverse physically plausible trajectories. During the rollout process, we compute the 3D point clouds ${{\rm PC_{w}} \in \mathbb{R}^{N\times3}}$ from the depth camera, where $\rm PC_{w}$ is represented under the world coordinate system (\emph{i.e.}, the center of the table) and $N$ is the number of points. 

\noindent \textbf{Coordinate transformation.}
As shown in Figure~\ref{fig:overview}, inspired by \cite{liu2022frame}, we propose to transform $\rm PC_{w}$ into the desired target coordinate system $\rm PC_{t}$, which makes the model more aware of the target position for control predictions. To capture dense interaction features between the robot and the object, we additionally transform $\rm PC_{w}$ into different robot joint coordinate systems (\emph{i.e.}, palm and finger tips). Finally, we combine point clouds representations under different coordinate systems ${\rm PC} \in \mathbb{R}^{N\times3(j+3)}$ and feed them into PointNet~\cite{qi2017pointnet}, where $j$ is the number of finger tips. Our visual policy predicts control commands based on extracted visual features and robot proprioceptive states. 

\noindent\textbf{Training.} We train our visual policy using either the behavior cloning (BC) or the recently proposed 3D diffusion policy~\cite{chi2023diffusion,ze2024dp3} from generated physically plausible trajectories. 
The BC model directly takes transformed point clouds and robot states as inputs and predicts robot actions. The diffusion policy employs the extracted 3D features from PointNet~\cite{qi2017pointnet} as the global condition for the denoising model and recovers actions from Gaussian noise. We train both models with $\ell 2$ loss between the predicted and ground-truth actions.
\section{Experiments}
\label{sec:expr}

We conduct extensive experiments and evaluate state-based and visual policies learned by our approach on three manipulation tasks for five seen and ten unseen objects.

\subsection{Experimental Setting}
\noindent \textbf{Video dataset.} 
The DexYCB~\cite{chao2021dexycb} dataset contains human demonstration videos of hand-object interactions. Following~\cite{qin2022dexmv}, we focus on five objects including \emph{mustard bottle}, \emph{tomato soup can}, \emph{sugar box}, \emph{large clamp} and \emph{mug}. 

For each object, we choose three videos which are most similar to those used in DexMV\footnote{The videos in DexMV are not publicly released.} in our experiments. 
Figure~\ref{fig:object_pose} presents motion retargeting results for these videos. Two evaluation protocols are used for benchmarking the performance of policies. Protocol \#1 trains the policy for each object separately and uses initial object poses from the first row of Figure~\ref{fig:object_pose}. For Protocol \#2, we train a unified policy for all five objects and adopt three poses for each object shown in Figure~\ref{fig:object_pose}. Furthermore, we evaluate the performance on ten unseen objects, namely \emph{master chef can}, \emph{tuna fish can}, \emph{pudding box}, \emph{gelatin box}, \emph{potted meat can}, \emph{banana}, \emph{pitcher base}, \emph{bleach cleanser}, \emph{wood block} and \emph{foam brick}.

\noindent \textbf{Simulation environment.} 
We follow previous works~\cite{rajeswaran2017learning,qin2022dexmv,wu2022learning} by using the  the Adroit robot hand and the MuJoCo simulator for fair comparison.
However, this setup is less realistic, as the robotic hand is not attached to an arm and can move freely.
To address this, we attach an Allegro robot hand to a UR5 arm, mirroring our actual hardware configuration. 
We use the SAPIEN~\cite{Xiang_2020_SAPIEN} simulator for this setup, which simplifies creating this more realistic environment and allows to speed up simulation. Unless specified otherwise, the Adroit hand in MuJoCo is used for benchmarking against state-of-the-art methods, while the Allegro hand in SAPIEN is employed for training visual policies and performing ablation studies.

\noindent \textbf{Manipulation tasks.}
We follow \cite{qin2022dexmv} to evaluate on three tasks.
The first \textbf{\emph{relocate}} task requires the robot to move an object to a target position. The \textbf{\emph{pour}} task aims to grasp a mug filled with particles and pour the particles into a container. The success rate is measured by the percentage of particles in the container. The  \textbf{\emph{place inside}} task aims to grasp a banana and place it into a mug. The success rate is measured by the percentage of the banana mesh in the mug.

\begin{figure}[t]
  \centering
\includegraphics[width=0.98\linewidth]{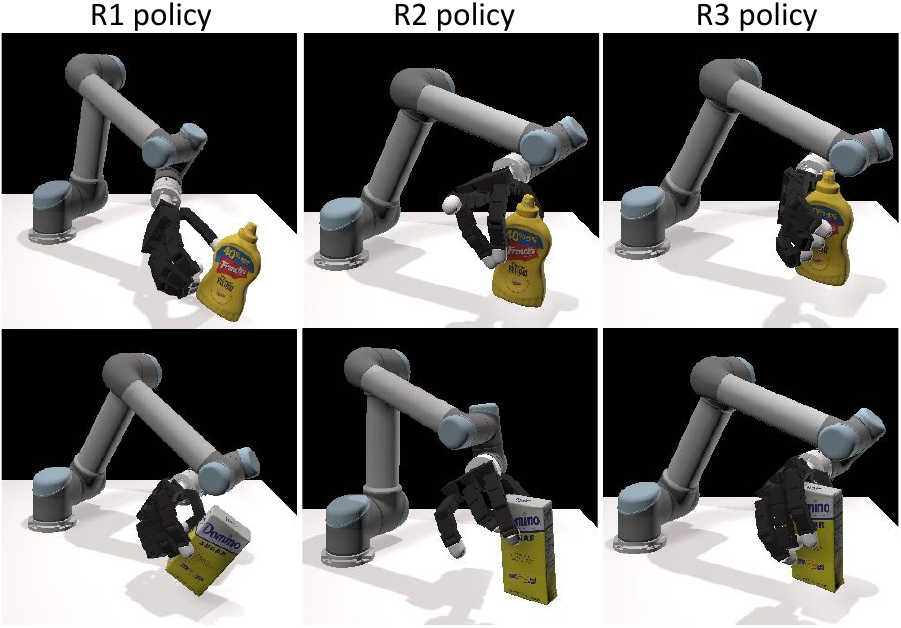}
\vspace{-0.2cm}
  \caption{Qualitative comparison of state-based policies using different rewards for Protocol \#1 and the Allegro hand. 
  R1 (w/o hand reward in pre-grasp) leads to unstable grasps. R2 (w/o hand reward in manipulation) results in unnatural hand actions. Our proposed approach R3 uses hand rewards at both stages and achieves the best performance.} 
  \label{fig:hand_reward}
  \vspace{-0.3cm}
\end{figure}

\begin{table}[t]
\centering
\caption{
Performance of our state-based policy using different hand rewards (Allegro hand, Protocol \#1). It follows the extracted human hand trajectory to manipulate objects (R1), reach the pre-grasp location (R2) or combine both (R3).}
\vspace{-0.1cm}
\setlength{\tabcolsep}{3pt}
\renewcommand\arraystretch{1.0}
\begin{tabular}{cccccccc}
\toprule
 & \multicolumn{2}{c}{Hand Reward}& \multirow{2}{*}{${\rm E_{o}}\downarrow$} & \multirow{2}{*}{${\rm E_{h}}\downarrow$} & \multirow{2}{*}{${\rm SR_{o}}\uparrow$} & \multirow{2}{*}{${\rm SR_{h}}\uparrow$} & \multirow{2}{*}{${\rm SR_3\uparrow}$} \\
   &Pre-grasp& Manipulation &  &  &  &  &  \\ \midrule
R1 &$\times$&\checkmark&0.048&0.21&0.35&0.00&0.00  \\
R2 &\checkmark&$\times$&0.0033&0.077&0.95&0.32&1.00   \\
R3 &\checkmark&\checkmark&\textbf{0.0019}&\textbf{0.032}&\textbf{0.97}&\textbf{0.79}&\textbf{1.00}\\
\bottomrule
\end{tabular}
\label{tab:hand_reward}
\vspace{-0.6cm}
\end{table}

\noindent \textbf{Evaluation metrics.} 
The success rate (${\rm SR}$) is the major metric for the above three tasks. For the \emph{relocate} task, we adopt the 10{\rm cm} threshold (${\rm SR_{10}}$) following \cite{qin2022dexmv,wu2023learning} to define the success rate. To measure the accuracy more precisely, we add a more rigorous threshold  of 3{\rm cm} and compute the success rate ${\rm SR_{3}}$. For the state-based policy, we additionally evaluate how well the physically plausible trajectory matches the reference trajectory using four metrics~\cite{dasari2023pgdm}:
${\rm E_{o}}$ computes the average object position error between the current object trajectory and its reference trajectory over time; 
${\rm E_{h}}$ computes the average finger tips position error between the current robot hand trajectory and its counterpart over time;
${\rm SR_{o}}$ reports the fraction of timesteps where ${\rm E_{o}}$ is below 1{\rm cm};
${\rm SR_{h}}$ computes the fraction of timesteps where ${\rm E_{h}}$ is lower than 5{\rm cm}. 

\smallskip
\noindent \textbf{Implementation details}. 
We use PPO~\cite{schulman2017proximal} to optimize the state-based policy. The RL training takes around 2 hours on a single A100 GPU for each video. 
We rollout 100 successful trajectories from the state-based policy for each video, where trajectories differ in initial object positions, orientations and target positions. During rollouts, we render 3D scene point clouds from the depth camera.
These data are used to train our visual policy, which takes robot joint positions and 3D point clouds as inputs.
Training our visual policy on a single A100 GPU takes around 10 hours using behavior cloning and about 20 hours using the diffusion policy, both based on data from 15 videos.
For testing, we run 300 episodes with different initial configurations and report the average success rate. The length of a single episode is 60, 80 and 100 for the \emph{relocate} task, \emph{place inside} task and \emph{pour} task, respectively. The real robot takes around 1 minute to execute an episode.

\subsection{Evaluation of the State-based Policy on the Relocate Task}
\label{sec:eval_state}
We evaluate the state-based policy on the \emph{relocate} task using both the Adroit and Allegro hands.
We follow Protocol \#1 to train the state-based policy then evaluate each policy on the same object with novel placement of the object. Specifically, we randomly change the initial object position and rotation around z-axis for each testing episodes.

\begin{table*}[ht]
\centering
\caption{
Comparison with state-of-the-art methods on \emph{relocate} task for state-based policies. 
We optimize a state-based policy using a single human video for each object (S6, S7, S8) (Protocol \#1), while previous methods usually need about 100 videos for each object (S2-S4). * indicates that we re-train the method from \cite{qin2022dexmv} with 20 DexYCB videos for each object.
}
\vspace{-0.1cm}
\setlength{\tabcolsep}{4pt}
\renewcommand\arraystretch{1.0}
\begin{tabular}{clccc|cc|cc|cc|cc|cc|cc}
\toprule
&\multirow{2}{*}{Methods} &\multirow{2}{*}{Robot Hands}  & \multirow{2}{*}{\begin{tabular}[c]{@{}c@{}}Training\\ algorithms\end{tabular}} & \multirow{2}{*}{\begin{tabular}[c]{@{}c@{}}Num.\\ videos\end{tabular}} & \multicolumn{2}{c|}{Mustard bottle} & \multicolumn{2}{c|}{Tomato can} & \multicolumn{2}{c|}{Sugar box} & \multicolumn{2}{c|}{Large clamp} & \multicolumn{2}{c|}{Mug} & \multicolumn{2}{c}{Avg.} \\
 & & &  & &${\rm SR_{10}}$&${\rm SR_3}$&${\rm SR_{10}}$&${\rm SR_3}$&${\rm SR_{10}}$&${\rm SR_3}$&${\rm SR_{10}}$&${\rm SR_3}$&${\rm SR_{10}}$&${\rm SR_3}$&${\rm SR_{10}}$&${\rm SR_3}$\\ \midrule
S1&DexMV~\cite{qin2022dexmv}&Adroit&TRPO~\cite{schulman2015trust}&0&0.06&-&0.67&-&0.00&-&0.51&-&0.49&-&0.35&-\\
S2&DexMV~\cite{qin2022dexmv}&Adroit&SOIL~\cite{Radosavovic2021}&97&0.33&-&0.98&-&0.67&-&0.89&-&0.71&-&0.72&-\\
S3&DexMV~\cite{qin2022dexmv}&Adroit&GAIL+~\cite{kang2018policy}&97&0.06&-&0.66&-&0.00&-&0.52&-&0.53&-&0.50&-\\
S4&DexMV~\cite{qin2022dexmv}&Adroit&DAPG~\cite{rajeswaran2017learning}&97&0.93&-&1.00&-&0.00&-&1.00&-&1.00&-&0.79&-\\
S5&DexMV~\cite{qin2022dexmv}$^{*}$&Adroit&DAPG~\cite{rajeswaran2017learning}&20&1.00&0.63&1.00&0.56&0.09&0.00&0.08&0.00&0.02&0.00&0.44&0.24\\ \midrule
S6&Ours&Adroit&PPO~\cite{schulman2017proximal}&1&\textbf{1.00}&\textbf{1.00}&\textbf{1.00}&\textbf{1.00}&\textbf{1.00}&\textbf{1.00}&\textbf{1.00}&\textbf{1.00}&\textbf{1.00}&\textbf{1.00}&\textbf{1.00}&\textbf{1.00}\\ 
S7&Ours&Allegro&PPO~\cite{schulman2017proximal}&1&\textbf{1.00}&\textbf{1.00}&\textbf{1.00}&\textbf{1.00}&\textbf{1.00}&\textbf{1.00}&\textbf{1.00}&\textbf{1.00}&\textbf{1.00}&\textbf{1.00}&\textbf{1.00}&\textbf{1.00}\\
S8&Ours w/o rot.&Allegro&PPO~\cite{schulman2017proximal}&1&0.85&0.85&1.00&0.99&0.99&0.97&0.98&0.95&0.97&0.94&0.96&0.94\\
\bottomrule
\end{tabular}
\label{tab:state_sota_relocate}
\vspace{-0.5cm}
\end{table*}
\begin{table}
\centering
\caption{Performance of visual policies for each object on \emph{relocate} task with ${\rm SR_3}$ using Protocol \#1 and Allegro hand.}
\vspace{-0.1cm}
\label{tab:visual_single_task}
\setlength{\tabcolsep}{3.5pt}
\renewcommand\arraystretch{1.0}
\begin{tabular}{ccccccccc}
\toprule
   & Algorithms   & \#Points & Mb.        & Tc.           & Sb.           & Lc.         & Mug           & Avg.           \\ \midrule
V1 & BC        & 512    & 0.99          & 0.82          & 0.82          & 0.78          & 0.90          & 0.86          \\
V2 & BC        & 2048   & 1.00          & \textbf{1.00}          & 0.87          & 0.91          & \textbf{1.00} & 0.96          \\
V3 & Diffusion & 2048   & \textbf{1.00} & 0.97 & \textbf{0.98} & \textbf{0.99} & 0.99          & \textbf{0.99} \\ \bottomrule
\end{tabular}
\vspace{-0.6cm}
\end{table}

\subsubsection{Ablation Studies}
We first perform ablations to validate the importance of our hand rewards at the pre-grasp and the manipulation stages in state-based policy learning. 
Table~\ref{tab:hand_reward} presents the averaged performance over all objects for policies with different reward functions for the Allegro hand. 
The policy in R1 uses the proposed hand reward $R_{m}^{h}$ in the manipulation stage but follows the previous reward function~\cite{qin2022dexmv,wu2023learning,rajeswaran2017learning} in the pre-grasp stage which is to minimize the distance between the robot and the object. 
It achieves poor performance due to the lack of guidance from the human hand trajectory to approach the object. As shown in the first column of Figure~\ref{fig:hand_reward}, the policy has difficulties in arriving at a plausible pre-grasp configuration to stably lift the object. 
The policy in R2 employs the proposed reward function $R_{p}$ in the pre-grasping stage but follows PGDM~\cite{dasari2023pgdm} (\emph{i.e.}, $R_{m}^{o}$ and $\mathds{1}_{\rm lift}$) in the manipulation stage. 
The reward function in PGDM constrains the object motions to be close to the reference trajectory but does not add any constraints on the hand motions, which results in unnatural robot actions as shown in the second column of Figure~\ref{fig:hand_reward}.
Our proposed policy in R3 utilizes hand rewards in both stages, which outperforms R1 and R2 on all the metrics and produces more robust and realistic grasps as shown in the last column of Figure~\ref{fig:hand_reward}.

\subsubsection{State-of-the-art comparison}
Table~\ref{tab:state_sota_relocate} presents the quantitative comparison of our state-based policy and state-of-the-art models~\cite{qin2022dexmv}. 
S1 to S4 denote four model variants reported in \cite{qin2022dexmv} for the Adroit hand. As their training data is not released, we re-train their best model DAPG~\cite{rajeswaran2017learning} (S4) using 20 DexYCB videos and report the results in S5 for fair comparison.
We can see that the performance drops significantly for some objects, \emph{i.e.}, \emph{large clamp} and \emph{mug}. 
This suggests that their approach needs a large number of videos to effectively learn how to handle objects with complex shapes.
Our proposed policy in S6 significantly outperforms all previous methods despite only using one training video per object. Though S4 also performs well for \emph{mustard bottle}, \emph{tomato soup can}, \emph{large clamp} and \emph{mug}, it easily knocks down thin objects (\emph{e.g.}, \emph{sugar box}) when the robot approaches the object and achieves poor performance. Our hand reward addresses this problem by mimicking the human pre-grasping trajectory. 
As shown in S7, we achieve the same performance with the Allegro hand. S8 is a variant of our model S7 without rotation augmentation in the z-axis during training. The performance only drops insignificantly  when tested for such rotations. This demonstrates the robustness of our method.

\subsection{Evaluation of the Visual Policy on the Relocate Task}
Here, we evaluate our proposed visual policy on the \emph{relocate} task and present a detailed experimental analysis.

\subsubsection{Learning from a single video per object}
We first investigate whether replacing the ground-truth object state in the state-based policy with visual point cloud inputs can affect the performance.
For each video, we follow Protocol \#1 to train a separate visual policy using the rollout data obtained from the state-based policy S7 in Table~\ref{tab:state_sota_relocate}, and evaluate under novel object placements as in the evaluation of the state-based policy.
The results are presented in Table~\ref{tab:visual_single_task}. Compared with V1 that uses 512 points, V2 samples 2048 points and improves the overall performance, which shows the importance of point clouds densities. By comparing V2 and V3, the 3D diffusion policy~\cite{ze2024dp3} can achieve more robust performance than the  plain behavior cloning algorithm. Our visual policies only perform slightly worse than our state-based policies.

\begin{table}
\centering
\caption{
Performance of unified visual policies on \emph{relocate} task under $\rm {SR}_3$ using Protocol \#2 and Allegro hand. We compare the performance using different number of human videos and point clouds under different coordinate systems.}
\vspace{-0.15cm}
\label{tab:visual_multi_task}
\footnotesize
\setlength{\tabcolsep}{2.4pt}
\renewcommand\arraystretch{1.0}
\begin{tabular}{cccccccccccc}
\hline
\multirow{2}{*}{} & \multirow{2}{*}{Algo.} & \multirow{2}{*}{Vid.} & \multicolumn{2}{c}{Coor. sys.} & \multicolumn{6}{c}{Seen} & \multirow{2}{*}{\begin{tabular}[c]{@{}c@{}}Unseen\\ avg.\end{tabular}}  \\ \cmidrule{6-11}
 &  &  & \multicolumn{1}{c}{target} & hand & Mb. &Tc.&Sb.&Lc.&Mug&Avg.&  \\ \midrule
V4 & BC & 1$\times$5 & \multicolumn{1}{c}{$\times$} &$\times$& 0.33 & 0.33 & 0.33 & 0.30 & 0.42 & 0.34 & 0.31  \\ 
V5 & BC & 2$\times$5 & \multicolumn{1}{c}{$\times$}&$\times$& 0.55 & 0.57 & 0.36 & 0.56 & 0.66 & 0.54 & 0.33 \\ 
V6 & BC & 3$\times$5 & \multicolumn{1}{c}{$\times$} & $\times$ & 0.69 & 0.80 & 0.83 & 0.86 & 0.88 & 0.81 & 0.38 \\ \midrule
V7 & BC & 3$\times$5 & \multicolumn{1}{c}{\checkmark}&$\times$& 0.93 & 0.92 & 0.99 & 0.93 & 0.99 & 0.95 & 0.37  \\ 
V8 & BC & 3$\times$5 & \multicolumn{1}{c}{\checkmark}&\checkmark& 0.93 & 0.97 & 0.97 & 0.98 & \textbf{1.00} & 0.97 & 0.41  \\ \midrule
V9 & Diff. & 3$\times$5 & \multicolumn{1}{c}{\checkmark} &\checkmark& \textbf{0.97} & \textbf{0.99} & \textbf{1.00} & \textbf{0.99} & 0.99 & \textbf{0.99} & \textbf{0.50} \\ \bottomrule
\end{tabular}
\vspace{-0.2cm}
\end{table}
\begin{table}[t]
\caption{Comparison with state-of-the-art methods on the \emph{pour} and the \emph{place inside} tasks using the Adroit hand. Our approach uses a single human video for each task, while previous methods need more than 91 human videos.}
\vspace{-0.15cm}
\centering
\setlength{\tabcolsep}{2pt}
\renewcommand\arraystretch{1.0}
\begin{tabular}{clcccc}
\toprule
\multirow{2}{*}{} & \multirow{2}{*}{Methods} & \multirow{2}{*}{\begin{tabular}[c]{@{}c@{}}Training\\ algorithms\end{tabular}} & \multirow{2}{*}{\begin{tabular}[c]{@{}c@{}}Num.\\ videos\end{tabular}} & \multirow{2}{*}{\begin{tabular}[c]{@{}c@{}}Pour\\ success rate\end{tabular}} & \multirow{2}{*}{\begin{tabular}[c]{@{}c@{}}Place inside\\ success rate\end{tabular}} \\ 
 &  &  &  &  \\ \midrule
L1&DexMV~\cite{qin2022dexmv}&TRPO~\cite{schulman2015trust}&0~/~0&0.01&0.03\\
L2&DexMV~\cite{qin2022dexmv}&SOIL~\cite{Radosavovic2021}&101~/~91&0.04&0.28\\
L3&DexMV~\cite{qin2022dexmv}&GAIL+~\cite{kang2018policy}&101~/~91&0.03&0.16\\
L4&DexMV~\cite{qin2022dexmv}&DAPG~\cite{rajeswaran2017learning}&101~/~91&0.27&0.31  \\ \midrule
L5&Ours&PPO~\cite{schulman2017proximal}&1~/~1&0.97&0.68  \\ 
L6&Ours& Diffusion&1~/~1&\textbf{0.97}&\textbf{0.68}  \\ \bottomrule
\end{tabular}
\label{tab:state_sota_pourplace}
\vspace{-0.7cm}
\end{table}

\subsubsection{Learning a unified multi-object visual policy from multiple videos}

Finally, we adopt Protocol \#2 and train a single policy for all objects with the Allegro hand and report its performance in Table~\ref{tab:visual_multi_task}. 
We test the policy under three initial poses for each object shown in Figure~\ref{fig:object_pose}. From V4 to V6, we gradually increase the number of videos for each object from one to three and observe a significant improvement in performance. Then,  V7 additionally transforms the 3D point clouds into the target coordinate system. As a result, the model becomes more aware of its target and largely improves the average performance for seen objects from 81\% to 95\%. To further incorporate fine-grained hand-object interaction features, we further transform 3D points into hand joint coordinate systems in V8. Benefiting from extracting rich hand-object interaction features, V8 achieves even better performance than V7. Different from V8, V9 trains the visual model using 3D diffusion policy and achieves an average success rate of 99\% under the challenging test scenario. 

To investigate the generalization of our visual policies, we evaluate their performance using three different initial poses for ten unseen YCB objects. In Figure~\ref{tab:visual_multi_task} from V4 to V6, our model better generalizes to novel objects when using more videos. V8 incorporates fine-grained interaction features and enhances generalization. Compared with the behavior cloning, V9 significantly improves the unseen success rate from 41\% to 50\% by using the 3D diffusion policy, which demonstrates better robustness and generalization abilities. 

\subsection{Evaluation of our policy on Pour and Place inside Tasks}
We further evaluate our approach on the \emph{pour} and \emph{place inside} tasks and report results in Table~\ref{tab:state_sota_pourplace}. 
Rows L1-L4 show results for state-based policies~\cite{qin2022dexmv} using the Adroit hand as well as 91 and 101 video demonstrations for \emph{pour} and \emph{place inside} tasks respectively. 
While we use only single human video for training, our state-based policy L5 for the Adroit hand achieves 97\% and 68\% success rates on these two tasks respectively, significantly outperforming 27\% and 31\% success rates corresponding to the best models of~\cite{qin2022dexmv}.
By taking advantage of high-quality trajectories generated by L5, our visual policy L6 also achieves high performance.

\begin{table}[]
\centering
\caption{Success rate of visual policies on \emph{relocate} task using the real robot. We evaluate 10 episodes for each object.}
\vspace{-0.15cm}
\setlength{\tabcolsep}{2.5pt}
\renewcommand\arraystretch{0.9}
\begin{tabular}{cccccccccc}
\toprule
\multirow{2}{*}{} & \multirow{2}{*}{Methods} &
  \multirow{2}{*}{\begin{tabular}[c]{@{}c@{}}Unified\\ policy\end{tabular}} &
  \multicolumn{6}{c}{Seen} & 
  \multirow{2}{*}{\begin{tabular}[c]{@{}c@{}}Unseen\\ Avg.\end{tabular}} \\ \cmidrule{4-9}
  &    &  & \multicolumn{1}{c}{Mb.}     & \multicolumn{1}{c}{Tc.}    & \multicolumn{1}{c}{Sb.}    & \multicolumn{1}{c}{Lc.}    & \multicolumn{1}{c}{Mug} & Avg. &  \\ \midrule
R1&BC    & $\times$ & 10/10 & 8/10 & 9/10 & 8/10 & 9/10&  0.88& -  \\
R2&BC    &\checkmark & 9/10 & 7/10 &7/10&5/10&8/10&0.72&0.58  \\ 
R3&Diffusion & \checkmark& 10/10&7/10&8/10&7/10&8/10&0.80&0.68\\ \bottomrule
\end{tabular}
\vspace{-0.3cm}
\label{tab:real_robot}
\end{table}

\begin{figure}[t]
  \centering
\includegraphics[width=0.86\linewidth]{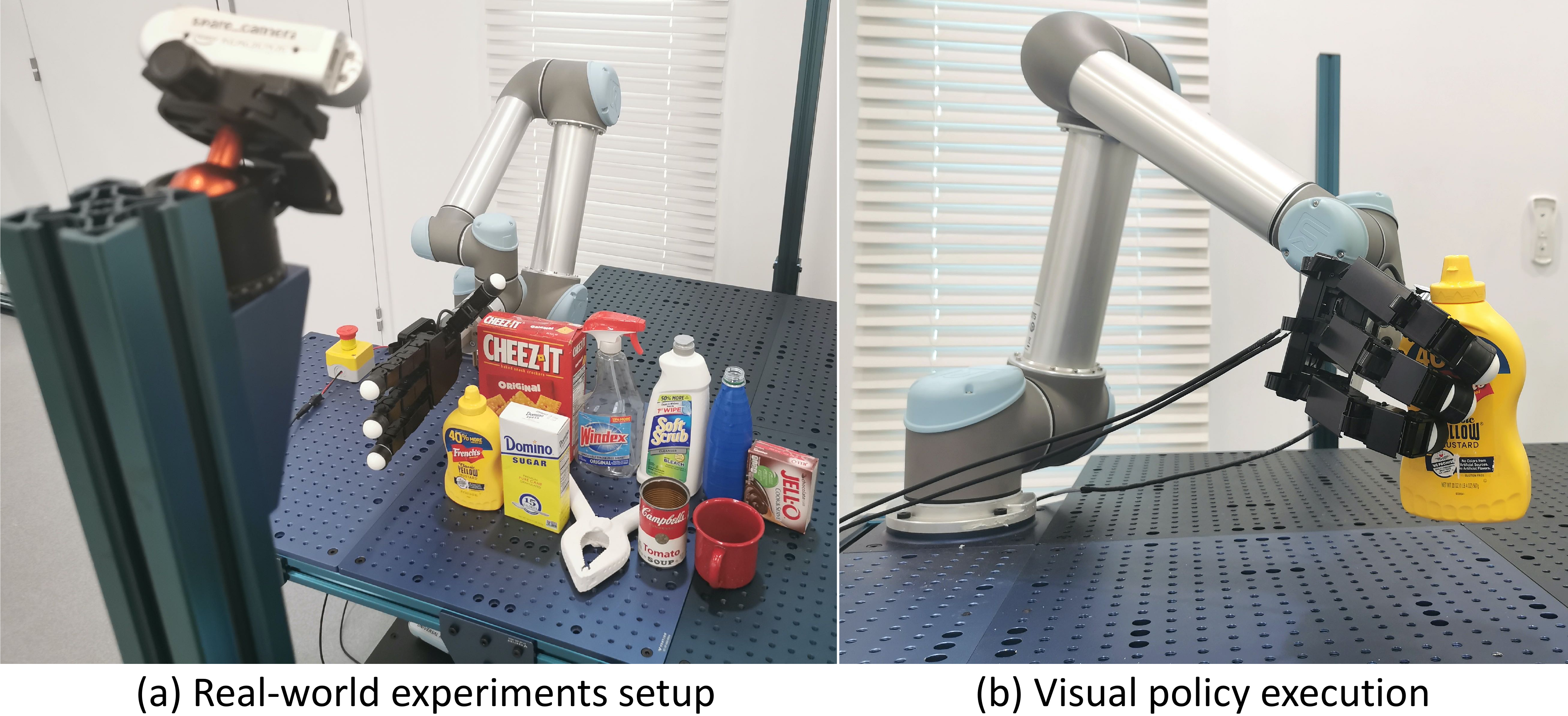}
\vspace{-0.15cm}
  \caption{Illustrations of our real-world robot experimental setup and the performance of our proposed ViViDex algorithm.} 
  \label{fig:real_robot}
  \vspace{-0.7cm}
\end{figure}

\subsection{Real world experiments}
To further demonstrate the advantages of ViViDex, we evaluate its performance on real-world dexterous manipulation.
As shown in Figure~\ref{fig:real_robot}(a), we use a UR5 robotic arm equipped with an Allegro hand and a single RealSense D435 RGB-D camera. Since point clouds for the real-world experiments are noisier than those in simulation, we use the state-based policy to collect data in the real world and then learn a unified visual policy based on real data. We run state-based policies in the simulator for specific object locations and run them on the real robot by placing the object in the same locations. This allows us to collect 3D scene point clouds and robot states for each execution step and build the real-robot training dataset.  We collect five robot trajectories for each object and train the visual policy using our collected real-robot data. As shown in Figure~\ref{fig:real_robot}(b), our visual policy can then manipulate the object. We summarize the quantitative results on the real robot in Table~\ref{tab:real_robot}. We adopt the initial pose for seen objects from the first row of Figure~\ref{fig:object_pose} and include five unseen objects for evaluation: \emph{cracker box}, \emph{spray bottle}, \emph{bleach cleanser}, \emph{water bottle} and \emph{pudding box}. R1 trains five policies separately and achieve an average success rate of 88\%. R2 and R3 learn unified policies for five objects and demonstrate the ability to grasp unseen objects. The diffusion policy R3 also achieves better performance than BC in real experiments.
\section{Conclusion}
\label{sec:conclusion}
We introduce ViViDex, a new framework for learning vision-based dexterous manipulation from human videos. Our approach extracts reference trajectories from human videos and uses them as a reward in training state-based policies with RL. This generates diverse physically plausible trajectories. We rollout these trajectories for training a visual policy, which takes as inputs robot proprioceptive states and 3D point clouds. To enhence visual policies, we further transforms 3D point clouds into hand-centered coordinate systems. 
We conduct extensive ablation experiments to validate the effectiveness of ViViDex on different simulators and a real robot. We show that our visual policies outperform state-of-the-art accuracy on different tasks by a significant margin and generalize to unseen objects. Future work aims at leveraging internet videos for acquiring more general dexterous manipulation skills and investigating advanced 3D pose estimation algorithms.

{
\renewcommand{\baselinestretch}{0.5}
\textbf{Acknowledgement}:
We thank Ricardo Garcia Pinel for help with the camera calibration, Yuzhe Qin for clarifications about DexMV and suggestions on the real robot depolyment. This work was granted access to the HPC resources of IDRIS under the allocation AD011013147 made by GENCI. 
It was funded in part by the French government under management of Agence Nationale de la Recherche as part of the “France 2030" program, reference ANR-23-IACL-0008 (PR[AI]RIE-PSAI projet), and the ANR project VideoPredict (ANR-21-FAI1-0002-01). Cordelia Schmid would like to acknowledge the support by the Körber European Science Prize.
}

\bibliographystyle{IEEEtran}
\bibliography{11_references}

\end{document}